\documentclass[pdflatex,sn-vancouver-ay]{sn-jnl}

\usepackage{graphicx}
\usepackage{multirow}
\usepackage{xcolor}
\usepackage{textcomp}
\usepackage{amsmath}
\usepackage{longtable}
\usepackage{geometry}
\usepackage{array}

\title{Federated learning, ethics, and the double black box problem in medical AI}

\author*[1,2]{\fnm{Joshua} \sur{Hatherley}}\email{jjh@hum.ku.dk}

\author[1,2,3]{\fnm{Anders} \sur{Søgaard}}

\author[4]{\fnm{Angela} \sur{Ballantyne}}

\author[5]{\fnm{Ruben} \sur{Pauwels}}

\affil[1]{\orgdiv{Center for the Philosophy of AI}, \orgname{University of Copenhagen, Denmark}}

\affil[2]{\orgdiv{Department of Communication}, \orgname{University of Copenhagen, Denmark}}

\affil[3]{\orgdiv{Department of Computer Science}, \orgname{University of Copenhagen, Denmark}}

\affil[4]{\orgdiv{Department of Primary Health Care and General Practice}, \orgname{University of Otago, New Zealand}}

\affil[5]{\orgdiv{Department of Dentistry and Oral Health}, \orgname{Aarhus University, Denmark}}

\abstract{Federated learning (FL) is a machine learning approach that allows multiple devices or institutions to collaboratively train a model without sharing their local data with a third-party.  FL is considered a promising way to address patient privacy concerns in medical artificial intelligence. The ethical risks of medical FL systems themselves, however, have thus far been underexamined. This paper aims to address this gap. We argue that medical FL presents a new variety of opacity -- \textit{federation opacity} -- that, in turn, generates a distinctive \textit{double} black box problem in healthcare AI. We highlight several instances in which the anticipated benefits of medical FL may be exaggerated, and conclude by highlighting key challenges that must be overcome to make FL ethically feasible in medicine.} 

\begin{document}

\maketitle

\section{Introduction}

Artificial intelligence (AI) is rapidly making its way into healthcare. The US Food and Drug Administration (FDA) has now granted regulatory approval to over 1,000 medical AI devices -- over 800 in the last five years alone. While clinical adoption remains in early stages, actual usage of certain leading devices is rapidly increasing \citep{wu2024characterizing}. Research in medical AI is also booming; the quality of studies has improved substantially in recent years, with more randomised controlled trials of medical AI systems underway than ever before. Ultimately, experts anticipate that AI will transform the practice of medicine, delivering substantial improvements to patient health, clinical efficiency, and the quality of patient care \citep{esteva2019guide,topol2019high}. 

But significant ethical concerns remain \citep{sparrow2019promise}. Training medical AI systems often requires hospitals to share their local patient data with third-party repositories or private technology corporations. Using medical AI also requires physicians to enter sensitive information into systems owned by third-party organisations with histories of poor regulatory compliance. And AI systems are notoriously vulnerable to malicious cyberattacks that can compromise patients' confidentiality and even threaten their safety. 

Enter federated learning (FL), a machine learning approach that allows multiple devices or institutions to collaboratively train a model without sharing their local data with a third party. A potential solution to longstanding privacy concerns, FL is increasingly mentioned in discussions of the future of healthcare AI \citep[see, for instance,][]{rajpurkar2022ai}. But the anticipated benefits of medical FL are not limited to its privacy preserving features. According to Rieke and coauthors, for example, FL has "significant potential for enabling precision medicine at largescale, leading to models that yield unbiased decisions, optimally reflect an individual’s physiology, and are sensitive to rare diseases” \citep[1]{rieke2020future}.

The ethical risks of medical FL systems themselves, however, have thus far been underexamined. This paper aims to address this gap. We argue that medical FL presents a new variety of opacity -- \textit{federation opacity} -- that in turn generates a new, \textit{double} black box problem in healthcare AI. We highlight several instances in which the anticipated benefits of medical FL appear exaggerated, and conclude by highlighting key challenges that must be overcome to make medical FL ethically feasible. As a secondary objective, this paper also provides an accessible overview of FL with the aim of bringing more researchers in ethics, philosophy, and the humanities into the discussion and debate.

The structure of the article is as follows. Section \ref{2} introduces FL, providing an overview of key concepts and processes. Section \ref{3} surveys recent research and applications of FL in healthcare, highlighting its anticipated benefits for patients and health systems. Section \ref{4} reviews FL's technical limitations and examines their impact on the technology's expected benefits. We suggest that some of these benefits -- specifically, those associated with data security, data use, model performance, and algorithmic bias -- may be exaggerated. Section \ref{5} introduces the double black box problem in medical FL, and analyses its implications for model security and performance, fairness and algorithmic bias, and explainability and accountability in medical AI. Section \ref{6} highlights some further ethical concerns relating to scalability and patient safety, data work and quality of care, and model updating and continual learning. Section \ref{7} summarises and concludes the article by calling on philosophers, ethicists, and other researchers in the medical humanities to play a stronger role in shaping the design, implementation, and use of this new wave of AI technologies in healthcare.

\section{What is federated learning?}\label{2}

FL "involves training statistical models over remote devices or siloed data centers, such as mobile phones or hospitals, while keeping data localized” \citep[50]{li2020federateda}. The first paper outlining the approach appeared less than ten years ago, in 2016 \citep{mcmahan2017communication}. Since then, most research in FL has appeared in online pre-print repositories rather than peer-reviewed journals. But this ratio is rapidly shifting as the approach gains mainstream popularity and approval \citep{pfitzner2021federated}. 

FL networks typically consist of two key components: training nodes and an aggregation server. \textit{Training nodes} refer to independent data sources or repositories that contribute their local data to train a global FL model without sharing this data beyond the firewalls fo the repository. In healthcare, training nodes can include hospitals and other healthcare providers. This is known as \textit{cross-silo FL}, in which a small number of trusted institutions contribute large, curated datasets with strict privacy constraints to train an FL model, along with substantial computational resources. In some cases, data can also be aggregated from smart technologies used by individuals (e.g. smart watches or toothbrushes). This known as \textit{cross-device FL}, in which a large number of edge devices contribute small data samples at frequent but irregular intervals with a high level of churn. The \textit{aggregation server}, in contrast, is responsible for coordinating the training process and for updating the global FL model by aggregating the local updates from each of the training nodes. 

A typical FL process proceeds in three stages \citep{nguyen2022federated}: 

\begin{enumerate}
    \item \textit{System initialisation and client selection:} The aggregation server selects training nodes to be involved in the next round of model training and updating. Training nodes can be selected according to various factors including data availability, computational resources, or prior performance metrics. 
    \item \textit{Distributed local training and updates:} The global model is sent to the selected training nodes. The training node then updates the gradient parameters of the global model using its own local data. Once all local updates are complete, the updated gradient parameters from each training node are sent back to the aggregation server. 
    \item \textit{Model aggregation and download:} The central server updates the global model by aggregating all the gradient parameters received from each training node. Several aggregation approaches are available. Federated Averaging (FedAvg), for instance, works by giving more weight to training nodes with larger datasets. Other aggregation approaches have been developed to address limitations in FedAvg, e.g., by compensating for data heterogeneity (FedProx) (see section \ref{5.2}) and local computational power (FedNova), or by using adaptive optimizers such as Adam (FedOpt) \citep{li2020federated, reddi2020adaptive, wang2020tackling}. The updated global model is then sent back out to each of the training nodes, and the process begins anew.
\end{enumerate}

There are three types of FL. The process just outlined illustrates \textit{horizontal FL}, which involves training a model on “horizontally split” data, or data that shares the same feature space but is sampled from different populations. For example, horizontal FL would be used in a network consists of hospitals that serve different patient populations but collect information about patients’ medical conditions using the same categories (e.g. both hospitals use the \textit{International Disease Classification} to categorise psychiatric conditions, rather than the \textit{Diagnostic and Statistical Manual of Mental Disorders}). In contrast, \textit{vertical FL} involves training a model on “vertically split” data, or data that is sampled from the same population but uses different features. For example, vertical FL would be used where the network consists of a hospital and a health insurance organisation that collect different classes of information about the same group of patients.  Finally, \textit{federated transfer learning} involves training a model on both horizontally and vertically split data. For example, federated transfer learning would be used in a network that consists of hospitals in different countries that each utilise distinct therapeutic programs.

Training nodes receive different types of models depending on the type of FL approach used. In horizontal FL, training nodes receive a shared global model that exhibits identical performance characteristics across all nodes. In vertical FL and federated transfer learning, however, training nodes receive distinct local models that each exhibit different performance characteristics depending on the node at which they are trained and used \citep{liu2024vertical}. Unlike horizontal FL, in other words, vertical FL and federated transfer learning result in \textit{synchronic variation} \citep{hatherley2023diachronic}.

\bigskip

\begin{longtable}{|>{\raggedright}p{6cm}|>{\raggedright}p{6cm}|}
    \hline \textbf{Key term} & \textbf{Definition}\\
    \hline \textit{Federated learning (FL)} & A machine learning approach that allows multiple devices or institutions to collaboratively train a model without sharing their local data with a third-party. \\
    \hline \textit{Training nodes} & Independent data sources or repositories that contribute their local data to training the global FL model. In healthcare, training nodes can include healthcare organisations (\textit{cross-silo FL}) or edge devices (\textit{cross-device FL}).\\
    \hline \textit{Aggregation server} & The central server responsible for: (a) coordinating local model training between training nodes; and (b) updating the global model by aggregating the gradient parameters from each of the training nodes.\\
    \hline \textit{Horizontal FL} & A FL approach used when training data is sampled from different populations but uses the same features (\textit{horizontally split data)}. Each training node receives an identical global model.\\
    \hline \textit{Vertical FL} & A FL approach used when training data is sampled from the same population but uses different features (\textit{vertically split data)}. Each training node receives distinct local models with different performance characteristics.\\
    \hline \textit{Federated transfer learning} & A FL approach used when training data is sampled from different populations and uses different features (\textit{horizontally} and \textit{vertically split data}). Each training node receives distinct local models with different performance characteristics.\\
    \hline
    \caption{Glossary of key terms.}
    \label{tab:glossary}
\end{longtable}

FL has drawn substantial attention for its potential benefits in medicine and medical AI. In the next section, we review the current state of FL in healthcare (section \ref{3.1}) and the benefits this technology is expected to bring for patients, physicians, and health systems at large (section \ref{3.2}). 

\section{Federated learning in healthcare}\label{3}

\subsection{Applications}\label{3.1}

FL is a relatively recent innovation in machine learning research. Already, however, it has been used to develop models that perform a variety of functions in clinical care. Over the past several years, for instance, FL systems have been used to:

\begin{itemize}
    \item screen for or diagnose COVID-19 \citep{peng2023evaluation,qayyum2022collaborative,soltan2024scalable}
    \item diagnose and predict Parkinson's disease and Alzheimer's disease \citep{chen2020fedhealth, danek2024federated, stripelis2024federated};
    \item diagnose skin, lung, prostate, and breast cancer \citep{alsalman2024federated,haggenmuller2024federated,rehman2024fedcscd}; 
    \item classify histopathology images, mammogram images, and pediatric chest x-rays \citep{adnan2022federated, kaissis2020secure, roth2020federated};
    \item perform brain tumor, whole-brain, pancreas, and tooth segmentation \citep{li2020multi, roy2019braintorrent, schneider2023federated, sheller2019multi, shen2021multi, wang2020automated};
    \item generate medical imaging reports \citep{chen2024medical};
    \item predict patients' next clinical visit \citep{ben2024federated}; 
    \item predict patients' risk of mortality and hospitalisation, along with their length of hospital stay \citep{brisimi2018federated, huang2019patient, lin2024ai};
    \item predict clinical outcomes for COVID-19 patients \citep{dayan2021federated};
    \item predict histological responses to neoadjuvant chemotherapy in triple-negative breast cancer patients \citep{ogier2023federated};
    \item predict patients' risk of sepsis and acute kidney injury \citep{pan2024adaptive}.
\end{itemize}

The impact of FL extends beyond clinical care. For population health and clinical research purposes, for instance, FL systems have been used to: 

\begin{itemize}
    \item benchmark the performance of medical AI systems across multiple clinical sites \citep{karargyris2023federated};
    \item detect depression from social media posts \citep{khalil2024federated};
    \item identify cortical similarities between patients with schizophrenia, major depressive disorder, and autism spectrum disorder \citep{rootes2024cortical};
    \item perform tensor factorization for computational phenotyping \citep{kim2017federated};
    \item discover new biomarkers from fMRI images \citep{li2020multi};
    \item assist in cohort construction for research studies through patient similarity learning \citep{lee2018privacy};
    \item improve drug discovery \citep{hanser2025data};
    \item assist in data interpretation for single and multi-site health studies \citep{sadilek2021privacy};
    \item detect and classify new variations of COVID-19 \citep{chourasia2023efficient}.
\end{itemize}

Several international projects in medical FL are also underway. For example, the Nordic FederatedHealth project aims to use FL to improve the detection of medical implants and adverse drug reactions in collaboration with hospitals and universities in Denmark, Sweden, Finland, Norway, and Estonia \citep{chomutare2024implementing}.\footnote{See \url{https://www.nordicinnovation.org/programs/federatedhealth-nordic-federated-health-data-network}.} The Federated Tumour Segmentation (FeTS) initiative also aims to improve tumour boundary detection using an FL approach consisting of data from 71 clients across 6 continents \citep{pati2022federated}. Moreover, the FeatureCloud project aims to develop a central platform for developing FL algorithms for clinical purposes and integrating them into European healthcare \citep{matschinske2023featurecloud}.

\subsection{Anticipated benefits}\label{3.2}

Proponents anticipate that FL could benefit patients, physicians, and health systems in a variety of ways. According to Yoo and coauthors, for example, FL “is a structural solution for existing data privacy violation problems of machine learning methods” in healthcare \citep[78]{yoo2022open}. Relatedly, \cite{rajpurkar2022ai} suggest that improved data security in FL systems can reduce the risk of patient privacy breaches. Indeed, Kaissis and coauthors even suggest that “federated learning approaches have arguably become the most widely used next-generation privacy-preservation technique, both in industry and medical AI application” \citep[308]{kaissis2020secure}.

But the anticipated benefits of medical FL are not limited to patient privacy. For FL is also praised for its potential to improve the performance and generalisability of medical AI systems, and to reduce the susceptibility of these systems to algorithmic biases. As Rieke and coauthors express, FL has “significant potential for enabling precision medicine at largescale, leading to models that yield unbiased decisions, optimally reflect an individual’s physiology, and are sensitive to rare diseases” \citep[1]{rieke2020future}. One reason for this is that FL approaches could substantially increase the volume and diversity of datasets used to train medical AI systems by avoiding the patient privacy concerns associated with data sharing. By doing so, FL could reduce AI developers' reliance on publicly available datasets in which certain patient groups (e.g. non-Europeans)\footnote{The particular demographic groups that are under- or over represented in training datasets can differ between medical specialties. While European patients with lighter skin are overrepresented in dermatological datasets, for instance, they are often underrepresented in dental training datasets due to tighter data protection regulations.} are often underrepresented \citep{van2024federated}. Finally, Ng and coauthors claim that FL "incorporates data from multiple institutions, thereby increasing its external validity. Such a model would be much more likely to generate accurate results, even for what may be an atypical patient at a certain hospital" \citep[854]{ng2021federated}.

FL is also praised for its potential to address challenges to the stability of medical AI systems over time. According to Chen and coauthors, for example, FL could be used to "address many of the cases of dataset shift and to mitigate disparate impact via model development on larger and more diverse patient populations" \citep[729]{chen2023algorithmic}. \textit{Dataset shift} occurs when a mismatch arises between the data on which a model was trained and the data on which it is used in practice. In healthcare, for example, patient populations can change over time (e.g. due to gentrification), which can degrade the performance of a model over time. FL has the potential to address dataset shift by enabling hospitals to continuously and collaboratively update their medical AI systems over time. As such, FL also has the potential to facilitate "continual learning" in which medical AI learn from new training data even after being deployed in clinical practice \citep{hatherley2023diachronic, sun2023federated}. As Rieke and coauthors observe, for instance, "combining the learning from many devices and applications, without revealing patient-specific information, can facilitate the continuous validation or improvement of [...] ML-based systems" \citep[4]{rieke2020future}

FL could also radically increase the scalability of medical AI systems. For as Ng and coauthors observe, FL

\begin{quote}
    brings about auto-scaling at almost no additional cost. When new hospitals participate, they bring more data and more computational resources. As the loop continues to run, an ever-enlarging dataset is fed to the model, while all computations continue to be made by the end-user. The global model is updated after users have trained their individual models, requiring minimal resources to aggregate models and thus making deployment much more economical \citep[854]{ng2021federated}.
\end{quote}

Finally, FL could address urgent sustainability issues associated with deep learning-based medical AI systems. In particular, growth in the complexity of medical AI is rapidly outpacing the rate of growth in computational power and consumption. However, Jia and coauthors suggest that FL could address these concerns "by offloading the computational workload from a single central server to multiple servers or to other edge devices with computing system development" \citep[693]{jia2023importance}. 

FL, therefore, is anticipated to deliver enormous improvements to the privacy, accuracy, fairness, adaptiveness, and sustainability of medical AI systems. But these soaring expectations for the future of medical FL must be brought back down to earth. While its virtues are undeniable, medical FL also suffers from a variety of significant limitations that challenge recent assessments of its anticipated benefits. In the next section, we discuss several of these limitations, focusing on data security (Section \ref{4.1}), data use (section \ref{4.2}), model performance (section \ref{4.3}), and algorithmic bias (section \ref{4.4}).

\section{Limitations}\label{4}

\subsection{Data security}\label{4.1}

FL is heralded as a structural solution to patient privacy concerns in medical AI. But this claim is somewhat misleading. Like other machine learning approaches, FL systems are vulnerable to cyberattacks that can be used to reveal sensitive patient health information. Membership inference attacks, for instance, can be used to access patient data that an FL model has "memorised" from its training set \citep{nasr2019comprehensive}. These attacks are particularly easy to perform on training nodes with smaller training sets, putting cross-device FL networks and smaller healthcare organisations at greater risk of patient privacy breaches \citep{bagdasaryan2019differential}. Membership inference attacks on FL systems are often constrained by limited data availability. However, generative adversarial network-based (GAN-based) reconstruction attacks can overcome this limitation by generating synthetic data to increase data diversity \citep{zhang2020gan}. GAN-based attacks are also difficult to identify, with existing tools often exhibiting significant delays in detection \citep{lai2022gan}.

Several security measures have been developed to protect against these attacks, including multi-party computation, homomorphic encryption, and differential privacy. However, they all require large amounts of computing power that exceed what many healthcare organisations have available. Homomorphic encryption, in particular, demands extensive computational resources to execute. Differential privacy approaches, moreover, tend to compromise model performance \citep{kairouz2021advances}, generate performance disparities between patient subgroups \citep{bagdasaryan2019differential}, or reduce the model's explainability \citep{rust2023differential}.

\subsection{Data use}\label{4.2}

Even if FL \textit{were} fully privacy preserving, hospitals may not be entitled to contribute patients' health data to train an FL model without their consent. Ethical issues remain, particularly with respect to data ownership and public trust. For example, patients may perceive themselves to be harmed when their data is used for purposes that are misaligned with their values or expectations, even where third-party data sharing has not occurred. Consider the following scenario:

\begin{quote}
    Suppose that James, a patient with a history of body dysmorphia, discovers his health data is being used to train a medical AI system to evaluate the attractiveness of rhinoplasty patients on a ten-point scale pre- versus post-surgery \citep[see][]{khetpal2022perceived}. James thinks that unattainable societal beauty standards contributed significantly to the onset of his psychological condition. And he feels horrified that his health information has been used to reinforce or perhaps even exacerbate these standards. Over the following week, James begins to fixate on the shape of his nose, and his self-esteem plummets after several months of stability. He contacts his doctor's office to complain and to revoke his consent for third party data sharing. However, the administrator informs him that consent is not necessary. After all, no third party data sharing has occurred since the AI system was developed using an FL approach. James feels trapped and exploited. For not only does he feel that his data being used against his will, he also feels that he is being made complicit in an endeavour that is fundamentally at odds with his values.
\end{quote}

Scenarios like this highlight that patient privacy is not all that matters in medical FL. Patients have relevant interests, and a perception of ownership, with respect to how and why their data is used; medical FL risks circumventing these interests.

Moreover, effective clinical care relies on patient trust to disclose intimate and personal details relating to mental and physical health. Trust is typically founded not on explicit contracts or promises from doctors, but on sociocultural norms, background regulations, and professional standards \citep{cruess2020professionalism}. But public trust in doctors has been failing, in part to the digitization and dehumanisation of health systems \citep{ansberry2025why, kitsios2022digital}. FL could contribute to this decline in public trust where patients feel surprised to discover that their health information has been used to train a medical FL system. The use of clinical data to train medical AI systems has generated controversy in the past and has potential threat to public trust in medicine. A paradigmatic example was the decision by the National Health Service in the United Kingdom to share 1.6 million medical records with Google’s DeepMind to develop a diagnostic app for kidney disease. In a subsequent review, UK Information Commissioner Elizabeth Denham argued that this was problematic because “patients would not have reasonably expected their information to have been used in this way” \citep[quoted in][]{hern2017royal}. In a similar vein, medical FL could threaten public trust where reasonable patients do not expect their data to be used to develop or improve these systems.  

Consent may, therefore, be necessary for hospitals to donate or for developers to use patients' data to train a medical FL system. On the other hand, when patients are made aware of the trade-offs associated with requiring researchers to ask for consent -- cost, time (for them and researchers), selection bias, and in some case impracticability \citep{laurijssen2022impractical} -- many are willing to tolerate less control over their data in exchange for research that produces public benefit and is fairly accessible \citep{ballantyne2022sharing, richter2019patient, street2014use}. Moreover, some argue that patients have a \textit{prima facie} obligation to participate in health research on the grounds they are beneficiaries of medical knowledge gained from prior research. To avoid the charge of "free-riding," therefore, patients should also contribute to the production of health knowledge \citep{ballantyne2018consent, porsdam2016facilitating}. Such an obligation might weaken the requirements for consent for secondary use of clinical data in cases where research is low risk, privacy is protected, in the public interest \citep{ballantyne2020public}.

\subsection{Model performance}\label{4.3}

FL is praised for its potential to substantially improve the performance and generalisability of medical AI systems. But these anticipated benefits may be less significant than pundits suggest. Compute limitations (discussed previously in section \ref{4.1}) pose a significant obstacle to local training. According to one recent systematic review, "in most papers reviewed, the local optimization of models required GPU compute compatibility attached to the data source, which is not currently found in most hospital environments" \citep[9]{zhang2024recent}. A variety of reproducibility challenges have also been identified in the medical FL literature. Only half of studies included in a recent systematic review, for instance, "followed the principles of reproducibility in either making their data publicly accessible and/or providing access to code or containers at time of publication” \citep[10]{crowson2022systematic}. Alarmingly few studies of medical FL have tested model performance using both holdout and validation cohorts thus far, despite this being standard practice in the machine learning community \citep{zhang2024recent}. 

Model performance and generalisability may also be compromised to compensate for other shortcomings of FL, including privacy concerns and communication inefficiencies. As discussed in section \ref{4.1}, for instance, data security measures such as differential privacy often come at the cost of model performance. Moreover, strategies for overcoming communication inefficiencies in FL networks (e.g. gradient compression, pruning, or knowledge distillation) often compromise model performance and generalisability \citep{shah2021model}. Further issues arise due to federation opacity and the double black box problem in medical FL (see section \ref{5.2}).

\subsection{Algorithmic bias}\label{4.4}

FL is credited with the potential to address, and even overcome, the problem of algorithmic bias in medical FL. However, FL may be less useful in addressing algorithmic biases than pundits anticipate. While FL approaches may increase the size of datasets on which models are trained, they cannot address many of the deeper systemic causes of algorithmic bias \citep{sauer2024federated}. Algorithmic biases often arise due to missing data in electronic health records, such as data from low health literacy patients who are less likely to utilise online patient portals or provide patient-reported outcomes \citep{gianfrancesco2018potential}. Algorithmic biases often also result from historical disparities in the quality of care between different patient cohorts. For example, Black patients tend to receive weaker pain relief than White patients despite comparable self-ratings of pain \citep{hoffman2016racial} and disparities in screening and testing for certain conditions often occur between different patient cohorts. Again, further issues also arise due to federation opacity and the double black box problem in medical FL (see section \ref{5.3}). Increasing the number of hospitals involved in training a medical AI system alone cannot address these issues. 

\bigskip

To conclude, FL suffers from a variety of limitations concerning data security and data use, model performance, and algorithmic bias. In several cases, these limitations cut directly against the benefits that FL is expected to deliver in healthcare. Caution is needed to avoid inflated expectations as to the actual capabilities and probable benefits of FL in healthcare. But FL also generates its own distinctive epistemic challenges and ethical risks in healthcare. Perhaps the most significant of these challenges arise from the fact that stakeholders cannot access or analyse the complete datasets on which medical FL models are trained, as we discuss in the following section.

\section{The double black box problem}\label{5}

Medical AI systems, particularly deep learning systems, are notorious for exhibiting \textit{inference opacity}, which occurs when users cannot determine why a model predicts a certain output (\textit{x}) as a result of certain inputs (\textit{a}, \textit{b}, \textit{c}) \citep{sogaard2023opacity}. Suppose a medical AI systems predicts that an outpatient (\textit{P1}) has an 82 percent likelihood of being readmitted to hospital within thirty days. Inference opacity is what precludes \textit{P1}'s physician from being able to understand the reasoning process taken by the model to arrive at this percentage. Inference opacity gives rise to the \textit{black box problem} in medical AI, which “occurs whenever the reasons why an AI decision-maker has arrived at its decision are not currently understandable to the patient or those involved in the patient’s care because the system itself is not understandable to either of these agents” \citep[767]{wadden2022defining}. The black box problem is thought to have a variety of concerning implications for patient safety, trust, shared decision-making, and responsibility \citep{bjerring2021artificial, hatherley2020limits, smith2021clinical}. 

\begin{figure}[h]
    \centering
    \includegraphics[scale=0.82]{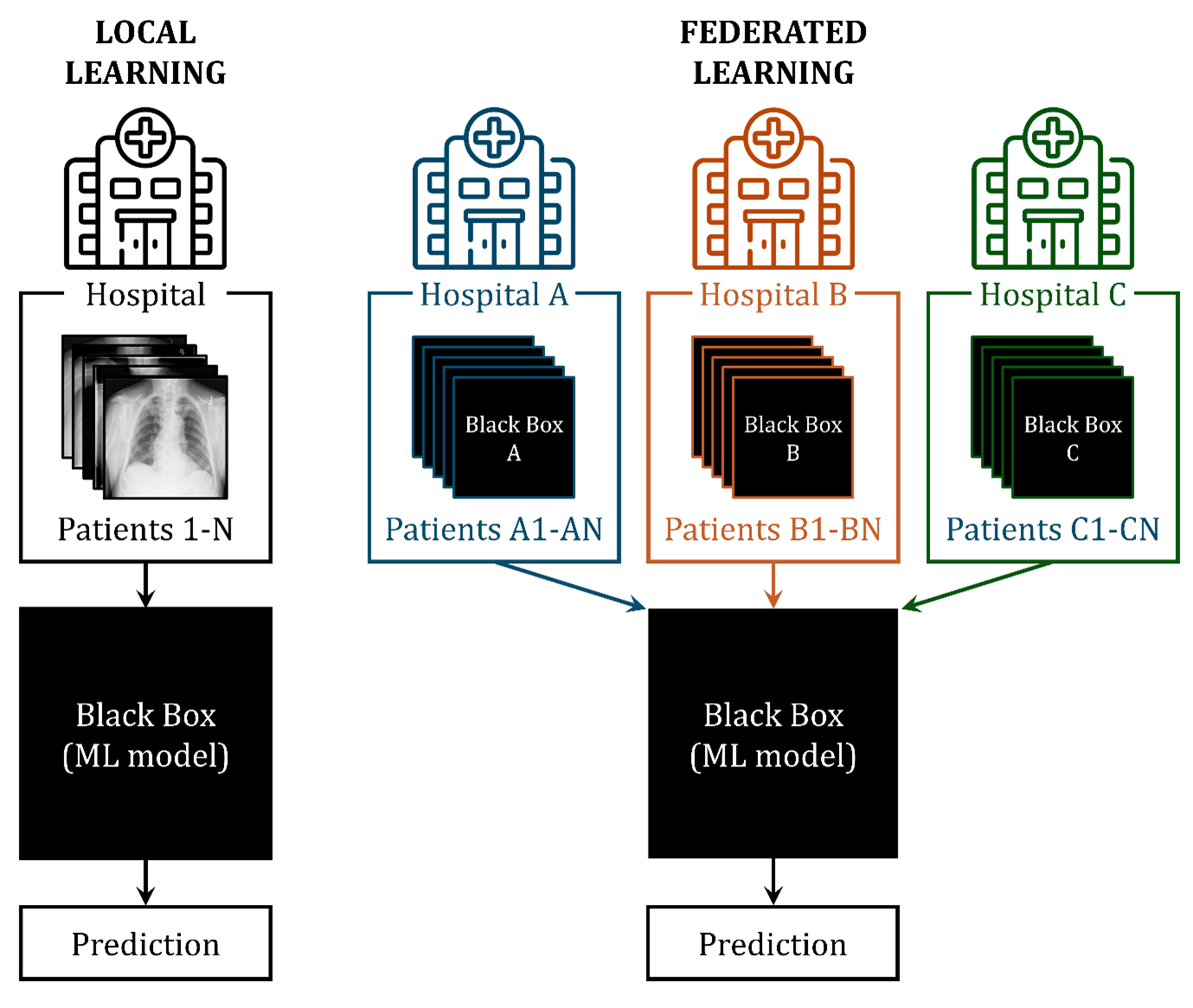}
    \caption{The double black box problem. Adapted from \cite{yoo2022open}.}
    \label{fig:dbbp}
\end{figure}

FL systems, like deep learning systems more generally, exhibit inference opacity. Suppose, for instance, that a doctor at training node \textit{n} uses a medical FL system trained on data from multiple hospitals to assist them in predicting patients' risk of developing breast cancer within 5 years. The doctor enters patient's (\textit{P2}) health data into the FL model, which outputs that \textit{P2} has a 73 percent likelihood of developing breast cancer within 5 years. \textit{P2} asks the doctor about the reasoning process that the FL system has taken to arrive at this conclusion, but the doctor cannot say for sure. In this respect, medical FL presents nothing new with respect to opacity and the black box problem. 

But FL also introduces a new and distinctive type of opacity into medicine and medical AI. For unlike deep learning systems, whose datasets are curated by their developers, FL approaches preclude developers from accessing or modifying the datasets on which FL models are trained. Each training node, of course, can access and curate the local data that they contribute to the FL process. They cannot, however, access or curate the local data of any of the other training nodes. Developers, moreover, cannot access or curate \textit{any} of the local datasets used to train an FL system. We refer to this as \textit{federation opacity}.

Federation opacity is epistemically significant because it gives rise to a \textit{double} black box problem in medical AI (see Figure \ref{fig:dbbp}). The \textit{double black box problem} occurs when:

\begin{enumerate}
    \item stakeholders cannot understand the reasoning process a model has taken to arrive at its outputs (\textit{inference opacity}); and 
    \item stakeholders cannot access, analyse, or curate the data on which a model has been trained (\textit{federation opacity}).
\end{enumerate}

Moreover, federation opacity presents a variety of distinctive ethical issues concerning model security (section \ref{5.1}), model performance (section \ref{5.2}), algorithmic bias (section \ref{5.3}), explainability (section \ref{5.4}), accountability (section \ref{5.5}), and fairness in medical AI (section \ref{5.6}).

\subsection{Model security}\label{5.1}

Federation opacity interferes with developers capacity to detect or protect against \textit{poisoning attacks}, a type of cyberattack that is used to degrade the performance of an FL model. Poisoning attacks can achieve this by corrupting the local data used to train an FL model at a specific node (\textit{data poisoning}). \textit{Adversarial attacks}, for example, are a type of poisoning attach that uses tampered data samples to cause intentional misclassifications, resulting in errors in the loss estimations propagated in future updates to the FL model \citep{kumar2023impact}. Adding structured ‘noise’ to an image enables adversarial attacks to compromise model performance while being visually undistinguishable from the original image. Access to model weights and inference results, as is the case in FL, also allows for highly effective ‘white-box’ attacks \citep[see][]{sheikh2024untargeted}. 

Poisoning attacks can also degrade the performance of an FL model by corrupting the local updates themselves (\textit{model poisoning}). For example, \textit{gradient manipulation} involves altering the local model gradients from one or more of the training nodes, interfering with the aggregation process and undermining the performance of the global model \citep{bagdasaryan2019differential}. Model poisoning attacks such as these have been found to have even stronger effects on model performance than standard data poisoning attacks \citep{fang2020local, shejwalkar2021manipulating}. 

The presence of federation opacity in FL systems makes them particularly vulnerable to poisoning attacks. Neither the aggregation server nor the developers can access the datasets on which an FL system is trained, which significantly restricts the strategies and opportunities available to developers for detecting or protecting against these attacks. For similar reasons, federation opacity interferes with developers' capacity to detect privacy breaches due to inference attacks, discussed in section \ref{4.1}. Despite this, few studies thus far have required the central server in an FL network to authenticate training nodes before receiving their updated weights or gradients \citep{zhang2024recent}. Some strategies for indirect detection have been developed, including robust aggregation and anomaly detection, yet they each carry significant limitations including costs to model robustness and scalability issues \citep{baruch2019little,jere2020taxonomy}. 

\subsection{Model performance (again)}\label{5.2}

Federation opacity interferes with developers' capacity to address performance and generalisability challenges associated with data heterogeneity in medical FL. Data heterogeneity issues occur due to heterogeneity in the training sets used to train an FL model. In medicine, for example, different hospitals use different databases, technical infrastructure, and clinical equipment. Take image data, for instance. In medical imaging, there are numerous makes and models for any given imaging modality. Over 100 different models of CT scanners are used in dentistry alone, and they all differ in contrast, noise, sharpness and (to an extent) artefacts. Moreover, data labeling quality can differ between (and even within) training nodes according to the data labelers' levels of expertise. At one hospital, for example, the data labeler may be a first-year resident, while another hospital may have multiple experienced people labeling data by consensus. 

Data heterogeneity not only interferes with the performance and generalisability of FL models \citep[see][]{babar2024investigating}; it can also lead to situations in which the global model converges on a solution that is biased against specific training nodes. Local updates from one training node that differ significantly from updates from other nodes in the network are likely to have little impact on the resulting global model. This presents fairness and equity concerns for both patients and hospitals who, despite contributing equally to the learning process of the global model, benefit less than other patients or hospitals in the federation (see also section \ref{5.5}). One way to address this challenge is to increase the influence of updates from sites with the highest loss during training to ensure certain nodes do not get "left behind" during training \citep{mohri2019agnostic}. However, this strategy carries the risk of “leveling down” the performance of the global FL model at higher performing client sites in order to equalise the performance of the global model with less influential client sites \citep[see][]{mittelstadt2023unfairness}.

Data heterogeneity presents substantial threats to the performance and generalisability of medical FL systems, yet federation opacity makes it extremely difficult to detect or remedy. For federation opacity precludes stakeholders from accessing or modifying the training sets used to train an FL model. Opportunities to identify or correct inconsistencies or variations in the quality of data labeling are extremely limited.  Developers may be unable to compensate for data heterogeneity without compromising model performance. Indeed, federation opacity also undermines developers' capacity to test different hyperparameter combinations on a given training dataset \citep{pfitzner2021federated}. The use of pre-training on public datasets, when available, can allow for some level of tuning, but more elaborate methods that operate within the FL process itself are currently unavailable. Standardising technical infrastructure or clinical equipment across participating institutions, moreover, is  both practically and financially unrealistic. Yet changes to clinical equipment or technical infrastructure may also need to be communicated to the developers of the FL system to reduce potential costs to performance, generalisability, or fairness in the system. 

\subsection{Algorithmic bias (again)}\label{5.3}

Federation opacity also undermines developers' capacity to directly identify or correct biases in local datasets that may be used to train an FL model. Developers would not be able to detect or redress, for example, instances in which the demographic composition of the overall dataset used to train an FL model is heavily skewed toward particular groups or demographic features. The same is also true for rarer medical conditions that may be underrepresented in a dataset. More data does not necessarily entail greater diversity. Where such limitations in a training dataset can in theory be addressed in a deep learning system, the same cannot be said for an FL model. 

FL is often praised for its potential to improve the detection of rare diseases. However, federation opacity presents substantial obstacles here. Since data is not shared between servers and institutions, there is no opportunity for developers to expand the number of rare events or atypical records, including records from underrepresented populations or patients with rare medical conditions. Studies of healthcare FL, moreover, do not – and cannot – provide any information about the data on which their models were trained, making it impossible to confirm that these datasets were sufficiently diverse and representative of the population that the model is intended to serve.

\subsection{Explainability}\label{5.4}

When federation opacity undermines our ability to identify or correct biases in local datasets, it is because federated learning by design prevents us from making inferences about these datasets. This also blocks training data attribution, e.g., through influence sketching \citep{Wojnowicz_2016}, influence functions \citep{koh2017understanding}, TraceIn \citep{pruthi2020estimating}, representer point selection \citep{yeh2018representer}, or related methods. In other words, federated learning, like differentially private training \citep{rust2023differential}, is at odds with this form of explainability. This is simply because you cannot have it both ways. Federated learning is designed to guarantee that you {\em cannot} make inferences about the training data from the inferences of the final model, whereas training data attribution is in general premised on your ability to do so. This does not mean federated learning is at odds with all explainability methods \citep[for an overview, see][]{sogaard2021explainable}. However, it means that federated learning limits our options considerably when we want to understand what influences the decisions made by the models we employ. We can explain these decisions in terms of input features and circuits in our networks, but {\em not} in terms of the training data that led to the model under scrutiny. 

\subsection{Accountability}\label{5.5}

Federation opacity also interferes with developers' capacity to identify training nodes responsible for errors or performance issues in an FL system.  Often, performance issues arise due to underlying issues in the dataset on which a model is trained. However, since the server cannot access the data from each individual site, it is incredibly difficult to identify how certain performance issues have come to be embedded in the global model. This is problem not only for performance and patient safety reasons, but also with respect to holding clients accountable for such errors (and their consequences). For example, suppose a client contributes low-quality data to an FL model, introducing spurious associations into the model’s reasoning that results in multiple patient mistreatments. Lacking traceability for these errors, it may be impossible to hold the offending client responsible for such mistreatments or deter other clients from contributing low-quality data in the future. The technical implications of untraceable errors are also immense. For example, if a server cannot determine which client is responsible for an issue, then the entire architecture of the model may need to be redesigned \citep{yoo2022open}. Redesigns of this sort could require developers to withdraw the global FL model from each site. In some cases, this could disrupt continuity of care and generate clinical and administrative inefficiencies.

\subsection{Fairness}\label{5.6}

Federation opacity makes it extremely challenging for developers to ensure all nodes contribute equitably to the federation, and to identify and eliminate “free riders” that benefit from the global model while contributing little. Where developers lack the ability to directly assess data contributions, nodes are relatively free to contribute poor quality data that could negatively impact the FL model's performance. Energy costs associated with training an FL model could also be distributed inequitably between training nodes and the aggregation server or between the nodes themselves.  In some cases, hospitals may also be forced to "pay double" for the privilege of partaking in an federation, “first for contributing data to train an FL model and then again to use the model” \citep[371]{bak2024federated}. Some strategies have been proposed to deal with these potential inequities. For example, incentive strategies have been developed that involve rewarding nodes for behaviour that aligns with the interests of the federation (e.g. by contributing more regularly or expeditiously to the learning process) \citep{tu2022incentive}. It is critical, however, that incentive-based strategies are not used to skew the performance of the FL model towards certain nodes, effectively punishing the patients at other nodes in the federation, and potentially for behaviour that they had no knowledge or control over. 

\bigskip

To summarise, medical FL presents a variety of distinctive epistemic and ethical challenges. Many of these challenges arise from the fact that stakeholders are unable to access or analyse the complete datasets on which medical FL models are trained. We have referred to this aspect of medical FL as federation opacity, and suggested that it generates a double black box problem in medical FL. But federation opacity and the double black box problem are not the only sources of medical FL's distinctive ethical challenges. In the next section, we discuss ethical risks pertaining to scalability (section \ref{6.1}), quality of care (section \ref{6.2}), and model updating and continual learning (section \ref{6.3}).

\section{Further ethical concerns}\label{6}

\subsection{Scalability}\label{6.1}

Scalability is highlighted as a key virtue of medical FL \citep{ng2021federated}. A single medical FL system could be deployed, not only across multiple hospitals, but multiple countries and even continents. But patient safety risks in medical AI increase proportionally to their scale. Errors in a system that is used to treat patients at one small rural clinic pose significantly lesser risk of patient harm compared to errors in a system that is used to treat patients in 10 busy city hospitals. If errors began to occur in a medical FL system utilised across, say, 50 hospitals in different countries spread over 4 continents, the results could be catastrophic - not only for patient health and safety, but for international relations and diplomacy.

\subsection{Quality of care}\label{6.2}

Medical FL is likely to have a significant impact on clinician workloads, yet this impact has thus far received little attention. FL, like other machine learning approaches, requires innumerably large datasets to function. But preparing training data for machine learning systems is can be costly and time-consuming. Thus far, many medical AI systems have been trained on largescale, publicly available datasets. In these cases, data prepartion tasks have often been outsourced to low-paid ghost workers with medical expertise through sites like Mechanical Turk. Outsourcing data work, however, is unlikely to be an option in FL. After all, patient data cannot be shared without consent beyond the institution in which it is stored. Physicians, therefore, will typically need to do the work involved in preparing training data (e.g. noise filtering, segmentation, and data labelling). This could substantially increase the already unmanageable day-to-day clerical work of physicians, a key factor in current rates of physician burnout and depression, and a significant threat to patient health and safety \citep{shanafelt2016relationship,sparrow2020high}.

Participating institutions may also need to make formal agreements for standardising data structures, annotations, and reporting protocols across different training nodes (see section \ref{5.2}). However, administrative and technical standardisation of this sort could increase the rigidity and inflexibility of healthcare administrative processes \citep[see][]{spencer2015brittleness}. This may also contribute to increased clinicians’ administrative workloads and decreased administrative or clinical efficiency, risking downstream costs to the quality of patient care.

Some researchers argue that FL could distribute the data work involved in medical AI developed more optimally between clinical sites. According to Ng and coauthors, for instance,

\begin{quote}
    the deployment of AI models requires periodic training and updating to remain current. This requirement may place an undue burden on radiologists, who must continually label a sufficient volume of studies necessary to retrain the model. At certain times when patient volume peaks, it may be difficult or even impossible for radiologists to produce enough annotated labels. However, because peak volume season may differ across hospitals, federated learning mitigates this issue, allowing radiologists at less busy hospitals to annotate studies while their counterparts at busier hospitals are too busy to do so \citep[854]{ng2021federated}. 
\end{quote}

This may, however, be overly optimistic. Clinicians' workdays are already overloaded with administrative tasks that reduce the amount of time they can spend on face-to-face patient care. Yet strong incentives are present in medical FL that could lead clinicians to prioritise data work over face-to-face clinical care to contribute as much data as possible to each training epoch in an FL network. Incentive mechanisms are, for instance, incorporated into the structure of medical FL to encourage training nodes to contribute large, high-quality datasets (as discussed in section \ref{5.6}). FedAvg, the most commonly used aggregation approach in medical FL, also weights the parameters and gradients from each training node according to the size of each node's local training set. The more training data the hospital contributes, therefore, the stronger its influence over the global model will be. As a result, managers or administrators may be tempted to put pressure on clinicians to prioritise performing datawork in their day-to-day responsibilities to realise the benefits associated with contributing more data to the FL network. After all, organisations tend to prioritise measurable outcomes over something as subtle and difficult to measure as "care" \citep{sparrow2020high}. Clinicians in the broader FL network, moreover, may not regularly be available to compensate for missing training data when other clinicians in the network are particularly busy. Whatever compensation that clinicians from other institutions could provide, moreover, does not directly compensate for missing datawork at a particular training node. For clinicians can only label patient data stored within their own organisations; data from other nodes in the network is unavailable to them. 

\subsection{Model updating and continual learning}\label{6.3}

Model updating and continual learning in medical FL present a variety of ethical concerns relating to patient safety, clinical interpretation, and informed consent, among others \citep{hatherley2023diachronic}. FL systems that engage in continual learning could, for instance, develop algorithmic biases even after being deployed in clinical practice. Such biases may not be detected before causing significant patient harm, particularly due to federation opacity. Continual learning in FL systems could also generate challenges for clinician-AI interaction. Model updating has been found to interfere with users’ mental models of AI systems, resulting in sudden drops in the performance of human-AI decision-making teams \citep{bansal2019updates}. Where medical FL systems undergo updates, the performance of physician-AI teams may be compromised, potentially leading to patient harm. This may be particularly likely where medical FL systems exhibit \textit{update opacity}, occurs when a user cannot understand why a model generates an output (\textit{x}) as a result of certain input data (\textit{a}, \textit{b}, \textit{c}), when a previous version of the model generated a different output (\textit{y}) on the basis of inputs \textit{a}, \textit{b}, and \textit{c} \citep{hatherley2025moving}.

In such scenarios, it is unclear how a doctor ought to proceed. The fact that the system has been updated between $t^1$ and $t^2$ provides no guarantee that the performance of the system has improved in this particular decision-making scenario, and that the clinician ought to accept the new output. Clinicians must be able to understand why an FL model generates different outputs at different points in time in order to collaborate responsibly with them. Several strategies for addressing update opacity have been proposed, including bifactual explanations, dynamic model reporting, and update compatibility. However, these strategies cannot overcome the problem \citep{hatherley2025moving}.

Further concerns may also be raised concerning the continual learning process in medicine. According to \cite{sparrow2024should}, for instance, continual learning medical AI systems may need to be classified (and therefore, regulated) as a form of clinical research. 

\section{Conclusion}\label{7}

Medical FL has significant potential to improve compliance of medical AI systems with ethical mandates and standards, particularly with respect to patient privacy and confidentiality. But it could easily transpire that current high hopes for the future of medical FL turn out to be exaggerated. Medical FL suffers a variety of limitations with respect to data security and use, model performance, and algorithmic bias that directly challenge several of the key benefits that this learning approach is anticipated to achieve in healthcare. Pundits should exhibit caution in their claims and expectations for what FL can realistically offer to patients, physicians, and health systems. Cautionary tales of overinflated expectations resulting in deep disappointment are evident in both the early and recent history of medical AI \citep{strickland2019ibm}. A clear vision for medical FL is necessary, but must be separated from overly optimistic expectations. 

A range of distinctive epistemic and ethical risks also arise from medical FL. Perhaps the most epistemically and ethically significant aspect of medical FL is that stakeholders cannot access or analyse the complete datasets on which medical FL models are trained and updated. We have called this aspect of medical FL "federation opacity". Federation opacity presents a broad range of obstacles, risks, and limitations with respect to model security and performance, fairness and algorithmic bias, and explainability and accountability in healthcare AI. A "double black box problem" also arises from it, since stakeholders are precluded not only from understanding the reasons underlying a model's outputs, but also from accessing or understanding the datasets on which the model is trained.

Medical FL has thus far evaded sustained attention from ethicists and humanities researchers with critical insights into how these systems ought to be designed, implemented, and used in healthcare settings. This article has aimed to draw philosophers' and researchers in the medical humanities' attention to the promise and perils of medical FL. Our hope is that this article stimulates further discussion, analysis, and debate as to the ethical trade-offs this technology presents in medicine. Philosophers and other researchers in the medical humanities have a critical role to play in shaping the future of healthcare AI, and medical FL offers a critical opportunity for them to execute this role. 

\bigskip

\backmatter

\bmhead{Acknowledgements} 

Thanks to Lauritz Aastrup Munch for comments on earlier versions of this paper. This work was funded by a Carlsberg Foundation Semper Ardens Advance (CF22-0243).

\bigskip

\bibliography{references}

\end{document}